\documentclass[letterpaper]{article} 
\usepackage{aaai2026}  
\usepackage{times}  
\usepackage{helvet}  
\usepackage{courier}  
\usepackage[hyphens]{url}  
\usepackage{graphicx} 
\usepackage{natbib}  
\usepackage{caption} 
\usepackage{algorithm}

\usepackage{amsmath}
\usepackage{booktabs}
\usepackage{amssymb}
\usepackage{pifont}  

\usepackage{subfigure}
\usepackage{enumitem}

\usepackage{multirow}
\usepackage{makecell}
\usepackage{amsthm}
\usepackage{xcolor}
\usepackage{algpseudocode}

\usepackage{newfloat}
\usepackage{listings}
\DeclareCaptionStyle{ruled}{labelfont=normalfont,labelsep=colon,strut=off} 
\floatstyle{ruled}
\newfloat{listing}{tb}{lst}{}
\floatname{listing}{Listing}
\title{De Novo Molecular Generation from Mass Spectra\\ via Many-Body Enhanced Diffusion}
\author{
    Xichen Sun\textsuperscript{\rm 1,3}\equalcontrib,
    Wentao Wei\textsuperscript{\rm 1,2}\equalcontrib,
    Jiahua Rao\textsuperscript{\rm 1}\correspondingauthor,
    Jiancong Xie\textsuperscript{\rm 1},
    Yuedong Yang\textsuperscript{\rm 1,4}\correspondingauthor
}
\affiliations{
    \textsuperscript{\rm 1}School of Computer Science and Engineering, Sun Yat-sen University, Guangzhou, Guangdong, China\\
    \textsuperscript{\rm 2}Pengcheng Laboratory, Shenzhen, Guangdong, China\\

    \textsuperscript{\rm 3}Shenzhen Loop Area Institute, Shenzhen, Guangdong, China\\ 
    
    \textsuperscript{\rm 4}Guangdong Provincial Key Laboratory of Computational Science, Sun Yat-sen University, Guangzhou, Guangdong, China

    \{sunxch7, weiwt8, xiejc3\}@mail2.sysu.edu.cn, \{raojh7, yangyd25\}@mail.sysu.edu.cn
}

\usepackage{bibentry}

\begin{document}

\maketitle

\begin{abstract}
Molecular structure generation from mass spectrometry is fundamental for understanding cellular metabolism and discovering novel compounds. Although tandem mass spectrometry (MS/MS) enables the high-throughput acquisition of fragment fingerprints, these spectra often reflect higher-order interactions involving the concerted cleavage of multiple atoms and bonds-crucial for resolving complex isomers and non-local fragmentation mechanisms.
However, most existing methods adopt atom-centric and pairwise interaction modeling, overlooking higher-order edge interactions and lacking the capacity to systematically capture essential many-body characteristics for structure generation.
To overcome these limitations, we present \textbf{MBGen}, a \textbf{M}any-\textbf{B}ody enhanced diffusion framework for de novo molecular structure \textbf{Gen}eration from mass spectra.
By integrating a many-body attention mechanism and higher-order edge modeling, MBGen comprehensively leverages the rich structural information encoded in MS/MS spectra, enabling accurate de novo generation and isomer differentiation for novel molecules. 
Experimental results on the NPLIB1 and MassSpecGym benchmarks demonstrate that MBGen achieves superior performance, with improvements of up to 230\% over state-of-the-art methods, highlighting the scientific value and practical utility of many-body modeling for mass spectrometry-based molecular generation. Further analysis and ablation studies show that our approach effectively captures higher-order interactions and exhibits enhanced sensitivity to complex isomeric and non-local fragmentation information. 
\end{abstract}

\begin{links}
    \link{Code}{https://github.com/biomed-AI/MBGen}
\end{links}

\section{Introduction}

The comprehensive understanding of cellular metabolism is essential for advancing basic biological research and applied biomedical sciences~\cite{metabolism1,metabolism2,rao2025multi,xie2024self}.
Metabolomics, which focuses on the systematic profiling of small molecules in biological samples, plays a pivotal role in revealing metabolic pathways and disease mechanisms~\cite{meta1,meta2,meta3,rao2024variational}. At the heart of metabolomics lies tandem mass spectrometry (MS/MS), a powerful analytical technique that enables high-throughput detection and structural elucidation of diverse metabolites through detailed fragmentation spectra~\cite{kind}. These spectra provide molecular fingerprints that enable the annotation of known compounds and the discovery of novel ones. The widespread adoption of MS/MS has markedly improved the resolution and depth of metabolomic analyses, propelling innovations in the field.

Despite its transformative impact, accurate and automated molecular structure generation from MS/MS data remains a fundamental challenge. The fragmentation spectra produced by tandem mass spectrometry are inherently complex, often reflecting not only the cleavage of individual chemical bonds but also higher-order interactions involving the concerted breakage of multiple atoms and bonds. These many-body effects encode rich structural information crucial for distinguishing complex isomers and interpreting non-local fragmentation mechanisms~\cite{isomers}.

 \begin{figure}[t]
  \centering
   \includegraphics[width=0.85\linewidth]{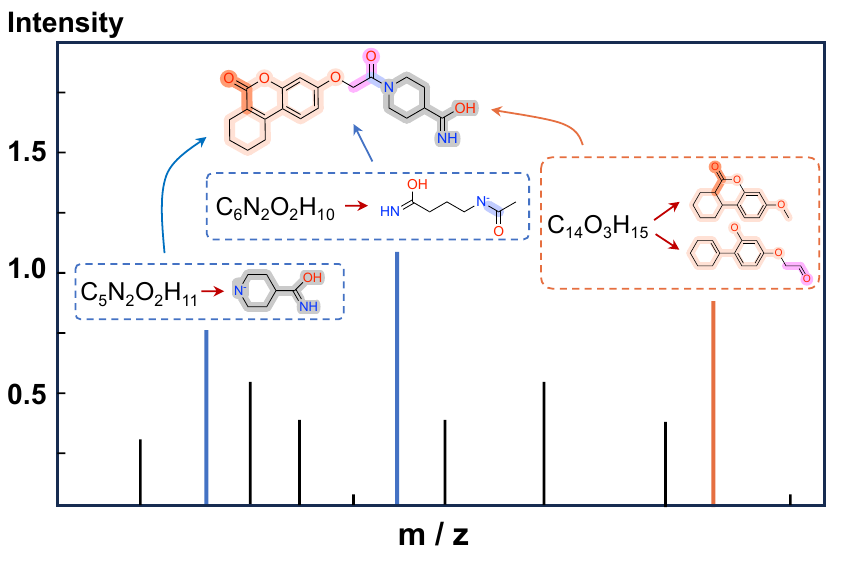}
   \caption{The mass spectrum comprises fragments from complex bond cleavages, where isomers within the same peak each contribute distinct yet essential insights for accurate molecular generation.
}
   \label{fig:intro}
\end{figure}
Existing computational approaches for molecular structure generation from MS/MS data primarily adopt an atom-centric perspective. For example, language models that map tokenized m/z values and intensities to SMILES strings~\cite{spec2mol,msnovelist,neuraldecipher}, as well as graph-based methods like MADGEN~\cite{madgen} and DiffMS~\cite{diffms}, represent and generate molecular structures by focusing on atoms as fundamental units. While these atom-centric strategies have improved prediction accuracy and enabled basic molecular assembly, they inherently overlook the rich chemical information encoded in bonds and the interactions between them. However, bond formation and cleavage events are central to the fragmentation processes observed in MS/MS spectra. Accurately modeling these processes requires a representation that directly captures the connectivity and dynamics of chemical bonds, rather than merely the arrangement of atoms.

Moreover, the fragmentation patterns observed in the MS/MS data often result from the concerted breaking of multiple bonds and complex, interdependent interactions that cannot be captured by simple pairwise modeling. Even advanced GNN-based models, such as those employing the Graph Transformer in DiffMS~\cite{diffms}, are limited in their ability to encode such higher-order relationships. Explicitly incorporating many-body interactions is therefore essential, as it enables the model to represent and predict multi-bond cleavage dynamics and non-local fragmentation mechanisms that are critical for faithful interpretation of MS/MS spectra. By capturing these complex interactions, the model can more effectively resolve structural isomers and generate chemically plausible de novo structures, ultimately leading to more accurate molecular generation from spectral data.

In this work, we present \textbf{MBGen}, a \textbf{M}any-\textbf{B}ody enhanced diffusion framework for de novo molecular \textbf{Gen}eration from mass spectra.
Specifically, MBGen differs from traditional atom-centric frameworks by adopting an edge-centric molecular generation strategy, modeling molecules at the level of chemical connectivity. This approach provides a more direct and chemically meaningful connection between spectral features and molecular structure, since bond formation and cleavage are fundamental to MS/MS fragmentation.  Furthermore, MBGen incorporates a many-body attention mechanism to explicitly capture higher-order interactions and concerted bond-breaking events. This allows the model to learn complex fragmentation pathways beyond simple pairwise or conventional GNN-based modeling, enabling more accurate interpretation of MS/MS spectra and effective differentiation of structural isomers. By integrating these chemical insights into a diffusion-based generative process, MBGen flexibly and reliably generates chemically plausible molecular structures from spectral data, advancing de novo molecular generation and isomer identification.

Experimental results on the NPLIB1~\cite{canopus} and MassSpecGym~\cite{massspecgym} benchmarks demonstrate that MBGen achieves superior performance, with improvements of up to 230\% over state-of-the-art methods.
MBGen consistently outperforms baseline models in distinguishing structural isomers, particularly in cases where isomeric species produce highly similar fragmentation patterns. 
Further analysis and ablation studies confirm that our approach effectively captures higher-order interactions and exhibits enhanced sensitivity to complex isomeric and non-local fragmentation information. These strengths position MBGen as a valuable tool for applications such as metabolite identification, drug discovery, and the structural elucidation of novel compounds in complex biological samples.
Our main contributions are as follows:
\begin{itemize}
\item We adopt an edge-centric molecular modeling strategy that directly represents chemical bonds and their connectivity, providing a more accurate foundation for interpreting MS/MS fragmentation.
\item We incorporate a many-body attention mechanism throughout the molecular generation process, explicitly capturing higher-order interactions and concerted bond-breaking events, which enables the model to better resolve complex fragmentation and structural isomers.
\item Extensive experiments demonstrate that MBGen significantly outperforms existing methods in both molecular structure generation accuracy and isomer differentiation, validating the value and practical utility of our approach.

\end{itemize}

\section{Related Work}
\subsection{Molecular Generation based on Mass Spectra}
Identifying molecular structures from mass spectrometry (MS) data remains challenging. Traditional approaches, such as those by~\citet{Markus} and tools like CSI:FingerID~\cite{CSI:FingerID}, predict molecular properties or fingerprints from tandem MS spectra and match them against molecular databases. However, these database-dependent workflows are computationally intensive and fundamentally limited by the coverage of reference databases, making the identification of novel compounds impossible.

To overcome these limitations, de novo molecular generation methods have been developed. These approaches predict molecular structures directly from MS data. For example, 
MSNovelist~\cite{msnovelist} combines predicted molecular fingerprints and formulas with an autoregressive model for molecule reconstruction.
Spec2Mol~\cite{spec2mol} employs an encoder-decoder framework, mapping spectra into a learned molecular embedding space for structure generation.
MADGEN~\cite{madgen} uses a two-stage process: scaffold retrieval from spectra, followed by conditional structure generation.
DiffMS~\cite{diffms} further advances the field with an end-to-end framework, pretraining spectra and structure modules separately before joint finetuning.
However, these methods typically adopt atom-centric and pairwise interaction modeling, overlooking higher-order edge interactions and lacking the capacity to capture essential many-body characteristics.

 \begin{figure*}[t]
  \centering
   \includegraphics[width=\linewidth]{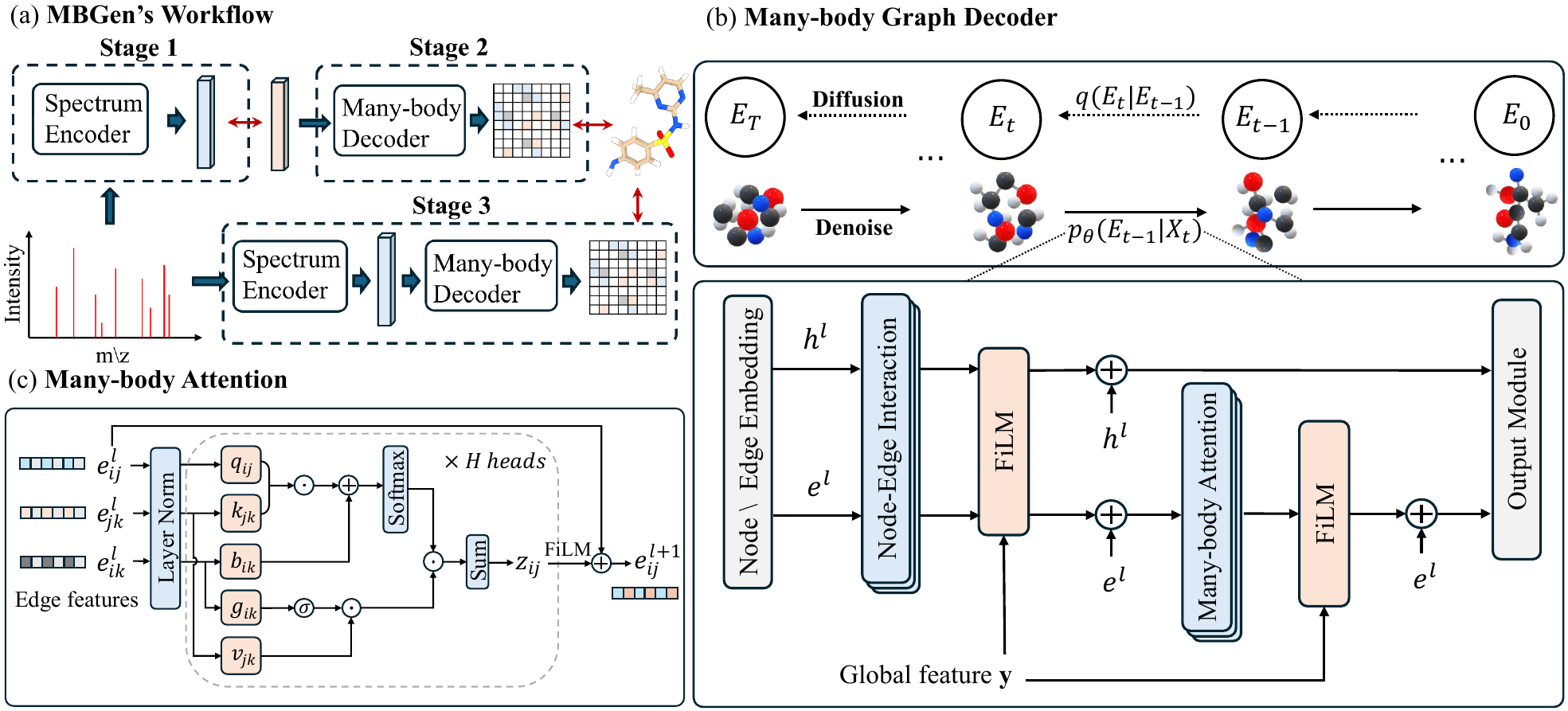}
   \caption{An overview of MBGen framework.}
   \label{fig:fig1}
\end{figure*}

\subsection{Many-body Interaction Modeling}
Recent advances in molecular modeling have highlighted the importance of many-body interactions for capturing complex dependencies between atoms in a molecule. While traditional geometric GNNs~\cite{graphormer, EGT, rao2025incorporating} mainly focus on pairwise interactions, limiting their expressiveness, recent architectures such as ViSNet~\cite{visnet}, GEM-2~\cite{gem2}, TGT~\cite{TGT}, and MABNet~\cite{MABNet} have incorporated higher-order mechanisms-modeling three-body and even four-body interactions-to enhance molecular representation learning. However, these approaches have largely been restricted to property prediction tasks, with limited exploration in molecular generation. 

In this work, we bridge this gap by introducing many-body interaction modeling directly into the generative process, enabling the decoder to capture richer geometric relationships and thereby improving the fidelity and accuracy of generated molecular structures.

\section{Methodology}

\subsection{Problem Formulation}

We formulate molecular structure generation from mass spectrometry (MS) data as a conditional graph generation problem. A molecule is represented as a graph \(G=(H,E)\), where
$H\in\mathbb{R}^{n\times d},E\in\{0,1\}^{n\times n\times k},$
with \(n\) heavy atoms, \(d\) feature dimensions, and \(k\) bond types.

Given a chemical formula which is obtainable from high‑resolution MS data using tools like SIRIUS~\cite{SIRUIS}, the node set \(H\) is fixed, and the task reduces to predicting \(E\) so that the resulting graph matches the observed spectrum.  The input spectrum \(S\) is comprised of \(m/z\) peaks and intensities \(I\), which are encoded into a vector \(y\) that provides global structural constraints.  Under this setting, our model learns to predict the adjacency matrix \(E\) conditioned on the fixed node features \(H\) and the spectral embedding \(y\).  

\subsection{Model Architecture}
Figure~\ref{fig:fig1} illustrates our overall framework. The input is a tandem mass spectrum, represented as a set of $(m/z, \text{intensity})$ peaks. The model first employs a spectrum encoder to extract a structural fingerprint from the spectrum. This fingerprint conditions a many-body graph diffusion decoder, which iteratively denoises a molecular graph to generate the final structure. The decoding process incorporates an edge-centric message passing mechanism and a many-body attention module to capture rich relationships and higher-order structural contexts. Our model adopts a three-stage training procedure, consisting of spectrum encoder pretraining, graph decoder pretraining, and end-to-end finetuning. 

\subsection{Spectrum Encoder}

Following previous work (DiffMS), we use the pretrained MIST~\cite{MIST} Formula Transformer as our spectrum encoder to extract molecular fingerprints from tandem mass spectra. Given an input spectrum $S = \{(m/z_i, \mathrm{intensity}_i)\}_{i=1}^N$, the encoder maps $S$ to a fixed-dimensional fingerprint vector $y$:
\begin{equation}
    y = \mathrm{Encoder}(S),
\end{equation}
where $y \in \mathbb{R}^d$ serves as the structural representation conditioning the subsequent molecular graph generation.

For the encoder, we first apply SIRIUS to annotate each peak with its most probable molecular formula, which is then concatenated with the corresponding peak intensity:
\begin{equation}
    x_i = [F_i ; \mathrm{intensity}_i] ,
\end{equation}
where $F_i$ embeds the molecular formula with a learned formula embedding function.

The set of peaks $X = \{x_i\}_{i=1}^{N}$ is then encoded by a Set Transformer comprising pairwise attention layers. Each layer updates peak embeddings by modeling their interactions as:
\begin{equation}
\mathrm{Attn}(x_i, x_j) = \frac{(Q_i + b_1) K_j + (Q_i + b_2) |F_i - F_j|}{\sqrt{d}},
\end{equation}
where $Q_i$, $K_j$ are the query and key vectors from learned projections of $x_i$, $x_j$, and $b_1$, $b_2$ are trainable bias terms. The value $|F_i - F_j|$ denotes the element-wise absolute difference between the embedded formulas of peaks $i$ and $j$.

Finally, the molecular fingerprint $y$ is obtained via mean pooling over the encoded representations:
\begin{equation}
    y = \frac{1}{N} \sum_{i=1}^N x_i .
\end{equation}

\subsection{Many-body Enhanced Graph Decoder}

\subsubsection{Edge-centric Molecular Modeling.}
For decoder, we adopt an edge-centric strategy, where information propagation and representation are centered on chemical bonds (edges) rather than solely on atoms (nodes). Instead of focusing exclusively on node-level embeddings, we explicitly construct edge features based on the associated node features and their relational context. 

We first initialize node features for each atom $i$ as:
\begin{equation}
    h_i^{(0)} = \mathrm{NodeEmb}(a_i),
\end{equation}
where $a_i$ encodes the atomic type and intrinsic properties.
Then, the pairwise (edge) embedding between nodes $i$ and $j$ is constructed as:
\begin{equation}
    e_{ij}^{(0)} = \mathrm{EdgeEmb}\big(h_i^{(0)}, h_j^{(0)}, r_{ij}\big),
\end{equation}
where $r_{ij}$ represents the relationship between $i$ and $j$.

In the edge-centric message passing, edge embeddings are updated in two sequential stages. First, node-edge interaction layers aggregate information from associated node features and the global feature $y$ to refine edge embeddings:
\begin{equation}
    e_{ij}^{(l+1)} = f_\mathrm{Edge}\Big(
        e_{ij}^{(l)},
        h_i^{(l)}, h_j^{(l)},y
    \Big).
\end{equation}

 Formally, at each layer $l$, the attention-weighted aggregation of node and edge features is computed as follows:
\begin{equation}
{\alpha}_{ij}^{(l)} =  \frac{(W_Q \cdot h_i^{(l)}) (W_K \cdot h_j^{(l)})^\top}{\sqrt{d}}  + W_E \cdot e_{ij}^{(l)},
\end{equation}
\begin{equation}
{o}_{ij}^{(l)} =  W_\alpha \cdot \alpha_{ij}^{(l)},
\end{equation}
where $W_Q$, $W_K$, $W_E$ and $W_\alpha$ are learnable projection matrices. 
Here, $o_{ij}^{(l)}$ represents the intermediate attention result integrating node and edge information.

Then, we apply a FiLM~\cite{film} mechanism to incorporate the global feature $y$ into edge representations. The modulation is applied as:
\begin{equation}
\mathrm{FiLM}(o_{ij}^{(l)}, y) =  y W_2 + o_{ij}^{(l)} \cdot (y W_1) + o_{ij}^{(l)},
\end{equation}
where $W_1, W_2$ are linear transformations.

Finally, the edge embedding is updated through a feed-forward network $\mathrm{FFN}_e$:
\begin{equation}
e_{ij}^{(l+1)} = \mathrm{FFN}_e\left(e_{ij}^{(l)} + \mathrm{FiLM}(o_{ij}^{(l)},y)\right).
\end{equation}

After updating edge features via node-edge interaction, to further capture higher-order chemical interactions, we incorporate a many-body attention mechanism, denoted as $f_\mathrm{ManyBody}$. 
Specifically, for each pair $(i, j)$, the many-body attention updates the pairwise embedding $e_{ij}$ by aggregating information not only from nodes $i$ and $j$, but also considering all neighbor pairs $(j, k)$ and their relationship with $(i, k)$, thus capturing interaction patterns among triplets $(i, j, k)$:
\begin{equation}
    e_{ij}^{(l+1)} = f_\mathrm{ManyBody}\Big(
    e_{ij}^{(l)},
    \big\{ e_{ik}^{(l)}, e_{jk}^{(l)} \mid k \in \mathcal{N}(j) \big\},
    y
    \Big),
\end{equation}
where $\mathcal{N}(j)$ denotes the neighborhood of node $j$. This enables the model to encode richer structural context by explicitly modeling interactions among triplets $(i, j, k)$, which is crucial for capturing complex fragmentation and higher-order chemical relationships during molecular generation.

\subsubsection{Many-body Attention Module.}

As shown in Figure~\ref{fig:fig1}(b), after the node-edge interaction, the pairwise embedding $e_{ij}$ is updated by a many-body attention module, which is further illustrated in detail in Figure~\ref{fig:fig1}(c). 
Specifically, at layer $l$, the many-body attention computes an intermediate output $\mathbf{z}_{ij}^{(l)}$ for each pair $(i, j)$ by performing a weighted sum over the value vectors of neighboring pairs:
\begin{equation}
\mathbf{z}_{ij}^{(l)} = \sum_{k=1}^N \alpha_{ijk}\, \mathbf{v}_{jk}
\end{equation}

where the attention weight $\alpha_{ijk}$ reflects the contribution of the neighbor pair $(j, k)$ when updating the target pair $(i, j)$, and $v_{jk}$ denotes the value vector of pair embedding $e_{jk}^{(l)}$. Specifically, $\alpha_{ijk}$ is computed as follows:
\begin{align}
s_{ijk} &= \frac{1}{\sqrt{d}}\, \mathbf{q}_{ij} \cdot \mathbf{k}_{jk} + b_{ik}, \\
\alpha_{ijk} &= \text{softmax}_k(s_{ijk}) \cdot \sigma(g_{ik}).
\end{align}

Here, the query, key, and value vectors are computed as linear projections of the corresponding pairwise embeddings:
\begin{equation}
\mathbf{q}_{ij} = W_Q \cdot e_{ij}^{(l)}  ,\mathbf{k}_{jk}  = W_K \cdot e_{jk}^{(l)}, \mathbf{v}_{jk}  = W_V \cdot e_{jk}^{(l)}, 
\end{equation}

and the bias term $b_{ik}$ and gating vector $g_{ik}$ are similarly computed from the third pairwise embedding ${e}_{ik}^{(l)}$:
\begin{equation}
{g}_{ik}  = W_G \cdot e_{ik}^{(l)}, {b}_{ik}  = W_B \cdot e_{ik}^{(l)},
\end{equation}
where all projection matrices $W_Q, W_K, W_V, W_G$ and $W_B$ are learned parameters.

The final attention weight is obtained by applying a softmax function and modulating it with a sigmoid gate $\sigma(g_{ik})$, which adaptively filters irrelevant interactions.

Similar to the previous step, we incorporate the global feature $y$ using the same FiLM method:
\begin{equation}
\mathrm{FiLM}(z_{ij}^{(l)}, y) =  y W_2 + z_{ij}^{(l)} \cdot (y W_1) + z_{ij}^{(l)}.
\end{equation}

The final updated embedding is then computed as:
\begin{equation}
e_{ij}^{(l+1)} = e_{ij}^{(l)} + \mathrm{FiLM}(z_{ij}^{(l)}, y) .
\end{equation}

This many-body update enables information flow along triplets without necessarily involving the junction node $j$, effectively alleviating the bottleneck and elevating the model's expressivity. A detailed
analysis of computational efficiency is provided in
Appendix E.


\subsection{Discrete Diffusion}

As shown in Figure~\ref{fig:fig1}(b), the discrete diffusion generation involves two processes:
$i)$ A diffusion process gradually corrupts the edge features of the molecular graph by introducing discrete noise;
$ii)$ A denoising process learns to reconstruct the molecular graph conditioned on spectral embeddings.

\subsubsection{Diffusion Process.}
Given a structure-spectrum pair $X = (G, y)$, where $G = (H, E)$ is a molecular graph and $E \in \mathbb{R}^{n\times n\times k}$ denotes the edge feature, we represent each edge feature $e$ as a k-dimensional one-hot vector, with class 0 being non-edge and classes 1 to k-1 corresponding to different bond types.
We model its diffusion process via a discrete forward process over time steps $t = 0, 1, \dots, T$, progressively adding noise to edge features using a categorical transition matrix $Q_t$:
\begin{equation}
    q(E_t | E_{t-1}) = E_{t-1} {Q}_t \quad \text{and} \quad q(E_t | E) = E \mathbf{\bar{Q}_t},
\end{equation}
where $ Q_t = \left[ q(e_t = j \mid e_{t-1} = i) \right]_{i,j = 0}^{k-1} $ and $\mathbf{\bar{Q}}_t = Q_t Q_{t-1} \cdots Q_1$.

For undirected graphs, noise is applied to the upper-triangular part of $E$, which is then symmetrized.

\subsubsection{Denoising Process.}
The reverse process begins with the fully corrupted edge matrix $ E_T$ and iteratively generates $ E_{t-1} $ from $ E_t $ until $ E_0 $, aiming to reconstruct the original edge types. A neural network $ f_\theta $ is trained to directly estimate $ E_0 $ from $ E_t $, conditioned on the noisy graph and the global feature. Consequently, the denoising transition from $ E_t $ to $ E_{t-1} $ can be expressed as:
\begin{equation}
q(E_{t-1} \mid E_t, E_0) \approx p_\theta(E_{t-1} \mid X_t),
\end{equation}
where $\theta $ serves as a learnable parameter.

\subsubsection{Modeling the Denoising Network $f_\theta$.}

To preserve the information of the molecular fragments interactions, we utilize the many-body attention mechanism to model $f_\theta$.
The denoising neural network \(f_\theta\)  is trained to reverse the corruption by predicting $\hat{p} = f_\theta(G_t, t, y)$.
$f_\theta$ takes a noisy graph $G_t$ as input and aims to predict the clean graph \(G\). To train \(f_\theta\), we optimize the cross-entropy loss L between the predicted probabilities \(\hat{p}\) and the true edge matrix \(E\):
\begin{equation}
    L =   \mathrm{CE}(\hat{p}, E) .
\end{equation}

After obtaining the trained network $ f_\theta $, new graphs are generated by estimating the reverse diffusion iterations $ p_\theta(E_{t-1} | X_t) $ , using marginalization over the network’s predicted distribution $\hat{p}$ :
\begin{equation}
     p_\theta(e_{ij}^{t-1} | X_t) = \sum_{e \in {E}} p_\theta(e_{ij}^{t-1} | e_{ij} = e, X_t) \hat{p}(e),
\end{equation}
where $p_\theta(e_{ij}^{t-1} | e_{ij} = e, X_t)={\rm max}\Big(q(e_{ij}^{t-1} | e_{ij} = e, e_{ij}^t),0\Big)$.
We can sample a discrete $ E_{t-1} $ from the distribution and then use it as the input for the denoising network in the next time step, iteratively performing the diffusion process until $ E_0 $.

\subsection{Training Paradigm}

\subsubsection{Spectrum Encoder Pretraining.}
Following the strategy of DiffMS~\cite{diffms}, we first pretrain the spectrum encoder on NPLIB1 and MassSpecGym datasets. Given an input spectrum $S$, the encoder is trained to predict the corresponding molecular fingerprint. This stage encourages the encoder to extract structural signals from spectral data and provides a strong starting point for subsequent finetuning. 

\subsubsection{Many-body Decoder Pretraining.}
To enhance the decoder's ability to generate molecular structures under strong structural constraints, we pretrain the decoder independently on a large collection of molecular fingerprint–structure pairs.
Specifically, instead of using spectrum embeddings as input, we directly use the molecular fingerprint as the structural condition $y$. The decoder is then trained to reconstruct the molecular graph, learning to generate true structures guided by the fingerprint feature.

\subsubsection{End-to-End Finetuning.}
After separate pretraining of the encoder and decoder, we jointly finetune the entire model using the same datasets as encoder pretraining. 
The encoder processes input spectra to produce the fingerprint embeddings $y$, which are then used to condition the many-body decoder.
In this stage, the model is trained to reconstruct the full molecular graph from scratch, aligning the predicted adjacency matrix with the ground-truth structure.

\begin{table*}[htbp]
\centering
\begin{tabular}{lcccccc}
\toprule
\multirow{2}{*}{\textbf{Model}} & \multicolumn{3}{c}{\textbf{Top-1}} & \multicolumn{3}{c}{\textbf{Top-10}} \\
\cmidrule(lr){2-4} \cmidrule(lr){5-7}
& \textbf{Accuracy~$\uparrow$} & \textbf{MCES~$\downarrow$} & \textbf{Tanimoto~$\uparrow$} & \textbf{Accuracy~$\uparrow$} & \textbf{MCES~$\downarrow$} & \textbf{Tanimoto~$\uparrow$} \\
\midrule
\multicolumn{7}{c}{\textbf{NPLIB1}} \\
\midrule
Spec2Mol & 0.00\% & 27.82 & 0.12 & 0.00\% & 23.13 & 0.16 \\
MIST + Neuraldecipher & 2.32\% & \ 12.11 & 0.35 & 6.11\% & \ 9.91 & 0.43 \\
MIST + MSNovelist & 5.40\% & 14.52 & 0.34 & 11.04\% & 10.23 & 0.44 \\
MADGEN & 2.10\% & 20.56 & 0.22 & 2.39\% & 12.69 & 0.27 \\
DiffMS & 8.34\% & 11.95 & 0.35 & 15.44\% & 9.23 & 0.47 \\
\textbf{MBGen} & \textbf{12.20\%} & \textbf{7.72} & \textbf{0.41} & \textbf{22.29\%} & \textbf{6.71} & \textbf{0.50} \\
\midrule
\multicolumn{7}{c}{\textbf{MassSpecGym}} \\
\midrule
SMILES Transformer & 0.00\% & 79.39 & 0.03 & 0.00\% & 52.13 & 0.10 \\
MIST + MSNovelist & 0.00\% & 45.55 & 0.06 & 0.00\% & 30.13 & 0.15 \\
SELFIES Transformer & 0.00\% & 38.88 & 0.08 & 0.00\% & 26.87 & 0.13 \\
Spec2Mol & 0.00\% & 37.76 & 0.12 & 0.00\% & 29.40 & 0.16 \\
MIST + Neuraldecipher & 0.00\% & 33.19 & 0.14 & 0.00\% & 31.89 & 0.16 \\
Random Chemical Generation & 0.00\% & 21.11 & 0.08 & 0.00\% & 18.26 & 0.11 \\
MADGEN & 1.31\% & 27.47 & 0.20 & 1.54\% & 16.84 & 0.26 \\
DiffMS & 2.30\% & 18.45 & 0.28 & 4.25\% & 14.73 & 0.39 \\
\textbf{MBGen} & \textbf{7.58\%} & \textbf{13.25} & \textbf{0.38} & \textbf{12.54\%} & \textbf{10.16} & \textbf{0.47} \\
\bottomrule
\end{tabular}
\caption{Evaluation of de novo molecular structure elucidation models on the NPLIB1~\cite{canopus} and MassSpecGym~\cite{massspecgym} datasets. The table presents top-1 and top-10 accuracy, Maximum Common Edge Substructure(MCES) scores, and Tanimoto similarity. \textbf{Bold} indicates the best performance. 
}
\label{tab:results}
\end{table*}

\section{Experiments}
\subsection{Experiment Setup}
In this section, we briefly describe the datasets, baseline methods, and evaluation metrics used in our experiments. Additional information can be found in Appendix A-C.

\begin{itemize}
\item \textbf{Datasets.}
We pretrain the decoder on a large scale of 2.8 million fingerprint–molecule pairs collected from DSSTox~\cite{dsstox}, HMDB~\cite{hmdb}, COCONUT~\cite{coconut}, and MOSES~\cite{moses}, covering diverse chemical structures. Evaluation is conducted on two public benchmarks: NPLIB1~\cite{canopus}, and MassSpecGym~\cite{massspecgym}.

\item \textbf{Baselines.}
We compare our method with several state-of-the-art approaches, including Spec2Mol~\cite{spec2mol}, MIST~\cite{MIST} combined with Neuraldecipher~\cite{neuraldecipher} or MSNovelist~\cite{msnovelist}, DiffMS~\cite{diffms}, and MADGEN~\cite{madgen}. 
We also reproduce the results of SMILES Transformer, SELFIES Transformer, and Random Chemical Generation from MassSpecGym.

\item \textbf{Evaluation Metrics.}
Following MassSpecGym, we report top-$k$ accuracy, Tanimoto similarity, and Maximum Common Edge Substructure (MCES) scores for $k=1$ and $k=10$. All models generate 100 molecular candidates for each spectrum. 

\end{itemize}

\subsection{Main results}
Table~\ref{tab:results} summarizes the main results of our method and representative baselines on the NPLIB1 and MassSpecGym datasets.
Our method achieves state-of-the-art performance across all metrics. On NPLIB1, our approach reaches a Top-1 accuracy of 12.20\%, surpassing the previous best result (DiffMS, 8.34\%) by a substantial margin. Similarly, on MassSpecGym, our method attains a Top-1 accuracy of 7.58\%, compared to 2.30\% for DiffMS. Significant improvements are also observed in Top-10 accuracy, MCES, and Tanimoto similarity, indicating that our model not only predicts more accurate structures but also generates candidates with higher substructural and overall molecular similarity.

Despite the use of chemical formula constraints, existing models such as DiffMS still exhibit relatively limited accuracy.
Our results demonstrate that explicitly modeling bond–bond (edge–edge) interactions, inspired by chemical fragmentation principles, can substantially improve both the accuracy and chemical plausibility of structure generation from MS/MS spectra. This highlights the importance of incorporating chemical knowledge into deep generative frameworks for molecular elucidation.

\begin{table}[t]
\centering
\normalsize
\setlength{\tabcolsep}{5pt} 
\begin{tabular}{@{}cc ccc@{}}
\toprule
\multicolumn{2}{c}{\textbf{Pretrain}} & \multicolumn{3}{c}{\textbf{Metrics}} \\
\cmidrule(r){1-2} \cmidrule(l){3-5}
\textbf{Enc.} & \textbf{Dec.} & \textbf{Accuracy} $\uparrow$ & \textbf{MCES} $\downarrow$ & \textbf{Tanimoto} $\uparrow$ \\
\midrule
\multicolumn{5}{c}{\textit{Top-1}} \\
\ding{55} & \ding{55} & 0.00\% & 17.96 & 0.17 \\
\checkmark & \ding{55} & 4.17\% & 15.73 & 0.26 \\
\ding{55} & \checkmark & 8.33\% & 15.20 & 0.30 \\
\checkmark & \checkmark & \textbf{12.20\%} & \textbf{7.72} & \textbf{0.41} \\
\midrule
\multicolumn{5}{c}{\textit{Top-10}} \\
\ding{55} & \ding{55} & 2.08\% & 14.26 & 0.25 \\
\checkmark & \ding{55} & 8.33\% & 13.25 & 0.36 \\
\ding{55} & \checkmark & 16.67\% & 11.65 & 0.44 \\
\checkmark & \checkmark & \textbf{22.29\%} & \textbf{6.71} & \textbf{0.50} \\
\bottomrule
\end{tabular}
\caption{Performance on the NPLIB1 dataset with and without pretraining of encoder and decoder.}
\label{tab:ablation}
\end{table}

\subsection{Ablation Study}
\begin{figure*}[t]
  \centering
   \includegraphics[width=\linewidth]{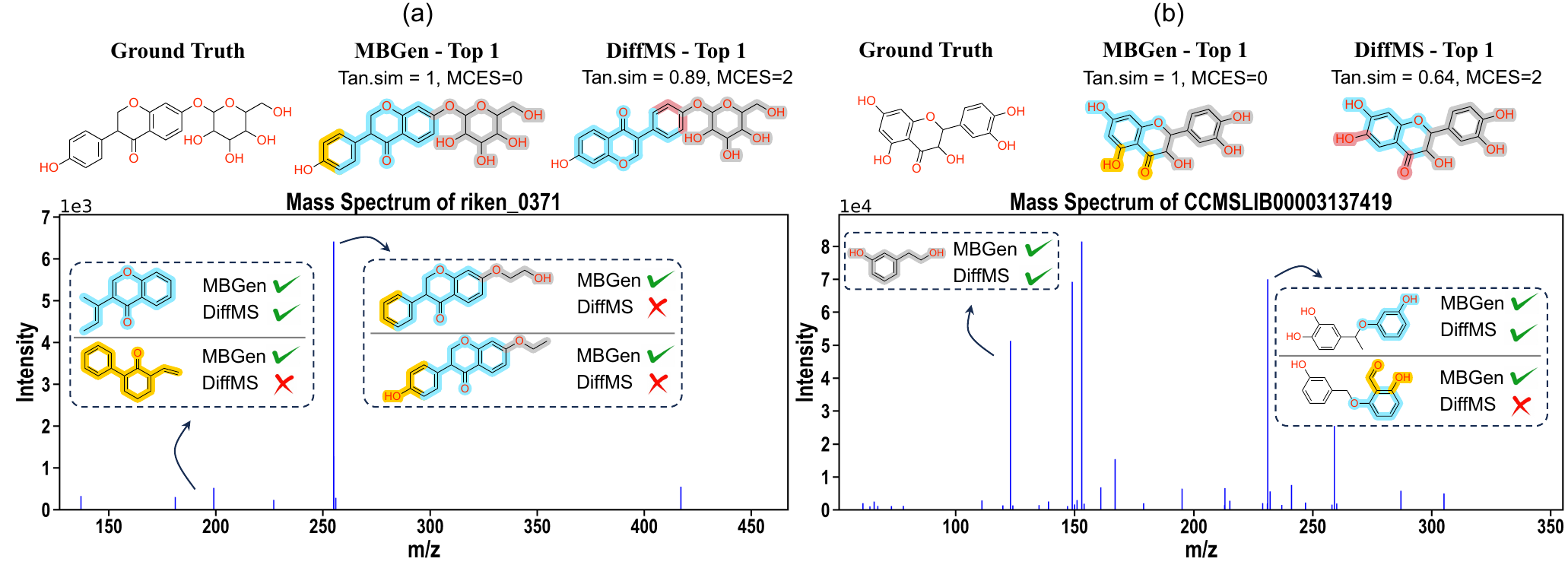}
   \caption{Case studies illustrating the superior performance of MBGen over DiffMS in de novo molecular structure generation, particularly in spectra featuring intra-peak isomers. }
   \label{fig:fig3}
\end{figure*}

\subsubsection{Many-body attention module.}
\begin{figure}[t]
    \centering
    \includegraphics[width=1\linewidth]{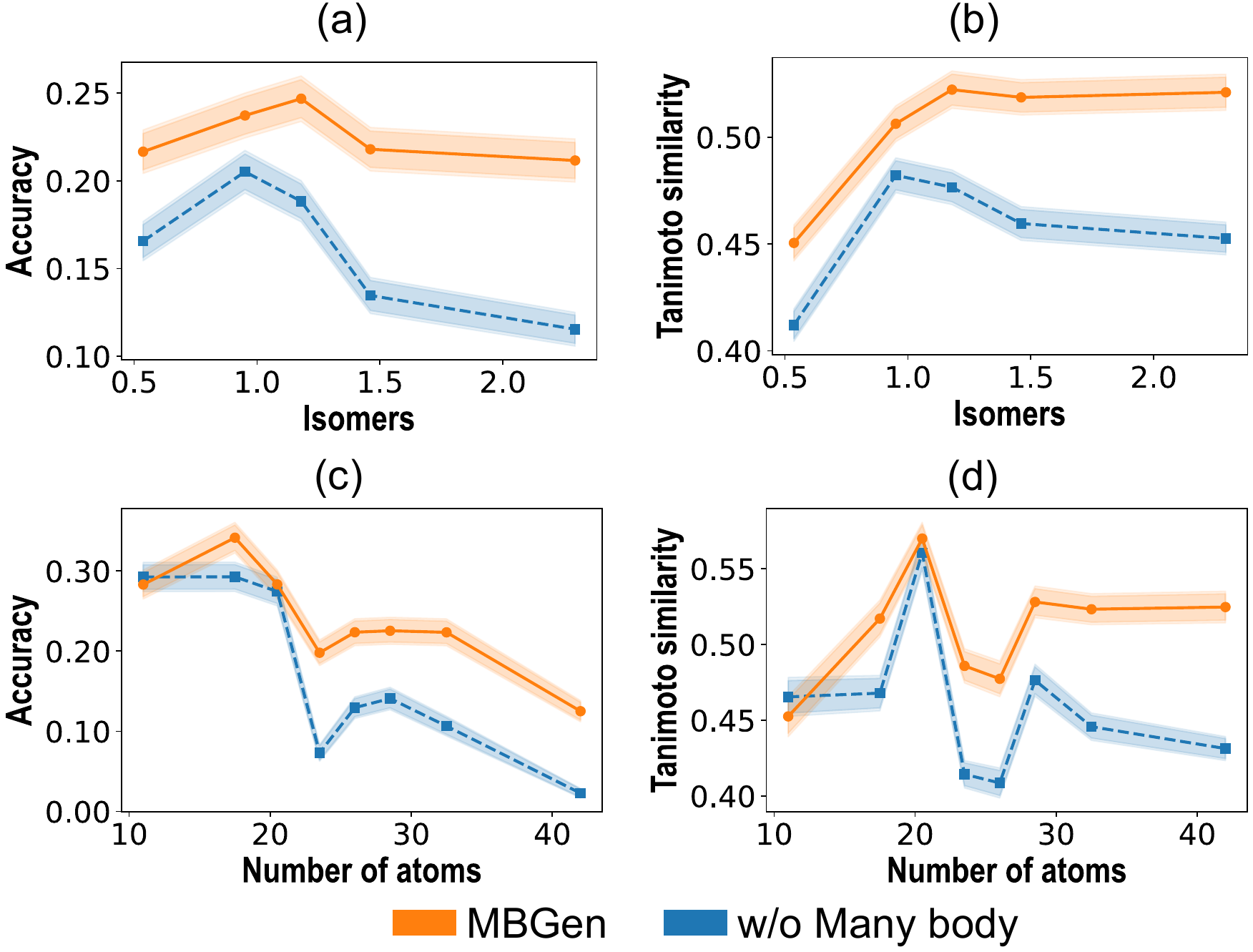}
    \caption{Ablation study on the NPLIB1 dataset, evaluating model performance across varying numbers of isomers and atoms.}
    \label{fig:fig2}
\end{figure}

To evaluate the efficacy of the many-body attention module, we conduct ablation experiments by removing it from the model (denoted as w/o MB). We use MAGMA to annotate fragment molecules for each peak in the mass spectra and compare model performance across varying numbers of isomers. As illustrated in Figure~\ref{fig:fig2}(a-b), both MBGen and the ablated model perform similarly when the average isomer count per peak is less than 1, but as the isomer complexity increases, MBGen maintains stable performance while the ablated model degrades, highlighting the module’s effectiveness.

We also assess the reconstruction of complex molecules (Fig~\ref{fig:fig2}(c-d)). For molecules with fewer than 23 atoms, both models perform comparably, but as the atom count increases, MBGen shows substantial gains. Notably, for molecules with over 40 atoms, MBGen achieves a Tanimoto similarity of about 0.525 versus 0.425 for the ablated model. Incorporating higher-order interactions thus enables robust modeling of complex molecules and offers deeper insights into molecular generation.

\subsubsection{Pretrain-finetune strategy.}
To assess the impact of our pretraining-finetuning strategy, we ablate the pretraining of the encoder and decoder components on the NPLIB1 dataset. As shown in Table~\ref{tab:ablation}, models without any pretraining yield poor performance, with Top-1 accuracy at 0.00\% and Tanimoto similarity at 0.17. Pretraining only the encoder or decoder provides moderate gains (e.g., Top-1 accuracy of 4.17\% and 8.33\%, respectively), while pretraining both achieves the best results: Top-1 accuracy of 12.20\%, MCES of 7.72 (lower is better), and Tanimoto of 0.41. Similar trends hold for Top-10 metrics, with the full strategy reaching 22.29\% accuracy, 6.71 MCES, and 0.50 Tanimoto. These results demonstrate that joint pretraining of encoder and decoder significantly enhances de novo generation by leveraging spectral and molecular priors.

\subsection{Case Study}
To demonstrate the interpretability and efficacy of MBGen, we present case studies highlighting its ability to capture isomer information within mass spectral peaks, leading to accurate de novo molecular structure generation where baselines like DiffMS falter. As shown in Figure~\ref{fig:fig3}(a), for the mass spectrum of riken-0371, both MBGen and DiffMS capture the chromone structural motif among isomers at the peak m/z 199.08. However, MBGen additionally identifies critical isomer fragments related to the benzene ring and chromone positional arrangements, which DiffMS misses. For the more complex isomer fragments at peak m/z 255.07, MBGen successfully captures them, whereas DiffMS fails, resulting in the loss of key structural insights.

Similarly, Figure~\ref{fig:fig3}(b) illustrates the spectrum of CCMSLIB00003137419, where MBGen accurately reconstructs the molecule by leveraging the many-body interaction to integrate complementary isomer details across peaks, whereas DiffMS produces an incorrect structure. This underscores how MBGen's many-body algorithm enhances interpretability by explicitly accounting for intra-peak isomer complexity, yielding superior predictions on challenging cases.

\section{Conclusion}
In this work, we propose MBGen, a many-body enhanced diffusion framework with edge-centric modeling for de novo molecular generation from mass spectra. We develop a pretraining-finetuning workflow incorporating an edge-enhanced transformer and a many-body attention module that leverages higher-order bond interactions and intra-peak isomer information, ensuring the model captures chemically nuanced representations from spectral data. We show that MBGen achieves state-of-the-art results across de novo generation benchmarks, and provide ablation studies and case analyses to demonstrate the effectiveness of our contributions and the potential to further enhance performance by scaling pretraining or integrating additional spectral priors.

\section{Acknowledgments}
This study has been supported by Shenzhen Medical Research Fund [C2403001], the Guangdong S\&T Program [2024B1111140001], the China Postdoctoral Science Foundation [2025M771540, GZB20250391], and the Lingang Laboratory [LGL-8888]. 

\bibliography{aaai2026}

\end{document}